\documentclass[runningheads]{llncs}
\usepackage[T1]{fontenc}
\usepackage{graphicx}
\usepackage{hyperref} 
\usepackage{color}
\usepackage{amsmath}
\usepackage{booktabs}
\usepackage{xcolor}
\usepackage{xspace,xpunctuate}

\usepackage{fancyhdr}
\newcommand{\ie}{{i.e.,}\xspace}

\newcommand{\eg}{{e.g.,}\xspace}

\begin{document}

\title{Your \emph{other} Left! \\Vision-Language Models Fail\\ to Identify Relative Positions in Medical Images} 

\titlerunning{Your \emph{other} Left!}

\author{Daniel Wolf\inst{1,2} \and
Heiko Hillenhagen\inst{2} \and
Billurvan Taskin\inst{2} \and
Alex Bäuerle\inst{3} \and
Meinrad Beer\inst{2} \and
Michael Götz\inst{2,*} \and
Timo Ropinski\inst{1,*} 
}

\authorrunning{D. Wolf et al.}

\institute{Visual Computing Group, Institute of Media Informatics, Ulm University, Germany  \and
Diagnostic and Interventional Radiology, Ulm University Medical Center, Germany \and
Axiom Bio, USA\\
\textsuperscript{*}\,\emph{Shared Last Authorship}\\[4pt]
\email{daniel.wolf@uni-ulm.de}
}

\maketitle      
\thispagestyle{fancy}

\begin{abstract}
Clinical decision-making relies heavily on understanding relative positions of anatomical structures and anomalies. Therefore, for Vision-Language Models (VLMs) to be applicable in clinical practice, the ability to accurately determine relative positions on medical images is a fundamental prerequisite. Despite its importance, this capability remains highly underexplored. To address this gap, we evaluate the ability of state-of-the-art VLMs, GPT-4o, Llama3.2, Pixtral, and JanusPro, and find that all models fail at this fundamental task. Inspired by successful approaches in computer vision, we investigate whether visual prompts, such as alphanumeric or colored markers placed on anatomical structures, can enhance performance. While these markers provide moderate improvements, results remain significantly lower on medical images compared to observations made on natural images. Our evaluations suggest that, in medical imaging, VLMs rely more on prior anatomical knowledge than on actual image content for answering relative position questions, often leading to incorrect conclusions. To facilitate further research in this area, we introduce the MIRP -- Medical Imaging Relative Positioning -- benchmark dataset, designed to systematically evaluate the capability to identify relative positions in medical images. Dataset and code are available on our \href{https://wolfda95.github.io/your_other_left/}{\textcolor{blue}{Project Page}}.

\end{abstract}

\section{Introduction}

\begin{figure}[h!]
    \centering
    \includegraphics[width=1\textwidth]{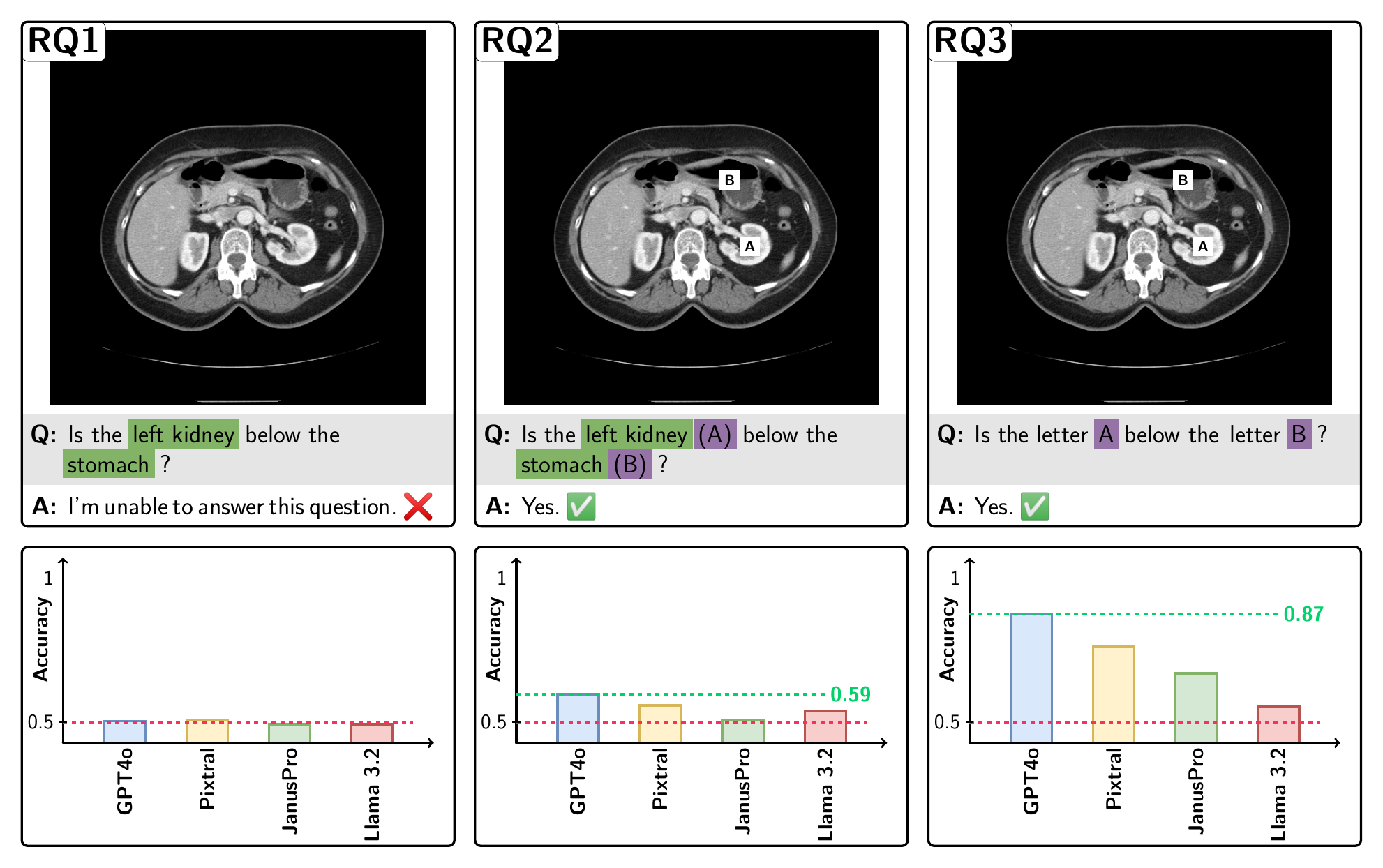}
    \caption{
    We compare four VLMs on their ability to determine relative positions of anatomical structures in medical images - a fundamental requirement for clinical applicability. All models fail when given plain images (RQ1), with only slight improvements when visual markers are introduced (RQ2). However, when anatomical names are removed from the question and models must rely solely on the markers, GPT-4o and Pixtral demonstrate remarkably improved accuracies (RQ3). 
    \textit{Note: The results show the best-performing marker type per model.}
    }
    \label{fig:overview}
\end{figure}

Imagine a radiology department where Vision-Language Models (VLMs) support complex tasks such as radiological report generation or surgical planning.
These systems could reduce diagnostic errors and improve patient outcomes~\cite{brady2017error,waite2017interpretive}.
While VLMs already achieve high accuracy in simple diagnostic tasks~\cite{almeida2023performance,srivastav2023chatgpt,jin2024hidden}, studies in computer vision show they struggle with spatial understanding, such as identifying relative positions of objects on images~\cite{rahmanzadehgervi2024vision,yang2023set,gao2025gllava}. This is problematic, as clinical decision-making relies heavily on understanding the spatial relationships between anatomical structures or anomalies~\cite{european2011good,datta2020understanding}. Localization mistakes by radiologists have led to serious consequences, including wrong-level spine surgeries~\cite{agolia2019preventing}, wrong-side procedures~\cite{joint2025sentinel}, or the failure to recognize the proximity of tumors to vessels~\cite{kang2021factors}.  
Just as understanding the spatial relationships is essential for radiologists, it is equally critical for VLMs to be clinically applicable. Without this capability, they cannot reliably describe localizations in reports.
To this end, our first research question (RQ1) is: \emph{“Can current top-tier VLMs accurately determine relative positions in radiological images?”}. We investigate this by focusing on a basic task: identifying the relative position of two anatomical structures in a computed tomography (CT) slice. Our findings indicate that even advanced VLMs, such as GPT-4o~\cite{hurst2024gpt} and Llama3.2~\cite{metallama}, fail at this fundamental task, raising concerns about their safe use in clinical routine.\\
\indent To address this limitation, we explore potential solutions. Previous work in computer vision has shown that visual markers, such as numbers or letters placed on objects, can enhance the capabilities of VLMs to determine relative object positions~\cite{yang2023set,cai2024vip}. This leads to our second research question (RQ2): \emph{``Can visual markers improve VLMs' ability to determine relative positions of anatomical structures in radiological images?''}. Since medical image segmentation is already well-established and highly accurate~\cite{wasserthal2023totalsegmentator,isensee2021nnu}, such markers on anatomical structures could, for example, be auto-placed based on segmentation model outputs.\\
\indent State-of-the-art VLMs already possess strong prior anatomical knowledge embedded within their language components~\cite{tarabanis2024performance}. In other words, they “know” where anatomical structures are typically located in standard human anatomy. We hypothesize that VLMs often base their answers on this prior knowledge rather than analyzing the actual image content. For example, when asked whether the liver is to the right of the stomach, a model might answer affirmatively without inspecting the image, relying solely on the learned norm that the liver is usually located to the right of the stomach. Such behavior could lead to critical misdiagnoses in cases where the actual positions deviate from typical anatomical patterns, such as in situs inversus, post-surgical alterations, or tumor displacement~\cite{eitler2022situs,kim2002postoperative,marcu2022impact}. If VLMs rely on language-based priors instead of visual evidence in these cases, the consequences could be severe. Similar issues have been observed in computer vision, where VLMs prioritize prior learned knowledge over actual image content~\cite{chen2024quantifying,vo2025vision}. Thus, our third research question (RQ3) is: \emph{“Do VLMs prioritize prior anatomical knowledge over visual input when determining relative positions in radiological images?”}\\
\indent To answer these questions, we introduce the \textbf{MIRP} (Medical Imaging Relative Positioning) benchmark dataset. Existing visual question-answering benchmarks on CT or MRI slices include anatomical and localization tasks, but do not focus directly on the fundamental task of determining relative positions~\cite{nan2024beyond,lau2018dataset,bae2024mimic}. Moreover, many localization tasks in these benchmarks can be solved using prior medical knowledge rather than actual image content, and an evaluation of this fallback is lacking. MIRP addresses this gap in three ways: through question-answer pairs assessing the relative position of two anatomical structures, by investigating whether visual markers improve model performance, and through random rotations and flips to prevent models from deriving correct answers based solely on learned anatomical norms. MIRP focuses on abdominal CT slices since these contain multiple anatomical structures per scan and are one of the most common radiological examinations~\cite{kanal2017us}.
We evaluate three leading open-source VLMs on MIRP: Meta’s Llama3.2~\cite{dubey2024llama,metallama}, Mistral’s Pixtral~\cite{mistralpixtral,agrawal2024pixtral}, and DeepSeek-AI's
JanusPro~\cite{chen2025janus,deepseekjanuspro}. We further include one of the top-performing closed-source VLMs currently available, OpenAI’s GPT-4o~\cite{hurst2024gpt}. The MIRP benchmark dataset, evaluation code, and all results are available on \url{https://wolfda95.github.io/your_other_left/}. 

\section{Methods}
\begin{figure}[h!]
    \centering
    \includegraphics[width=1\textwidth]{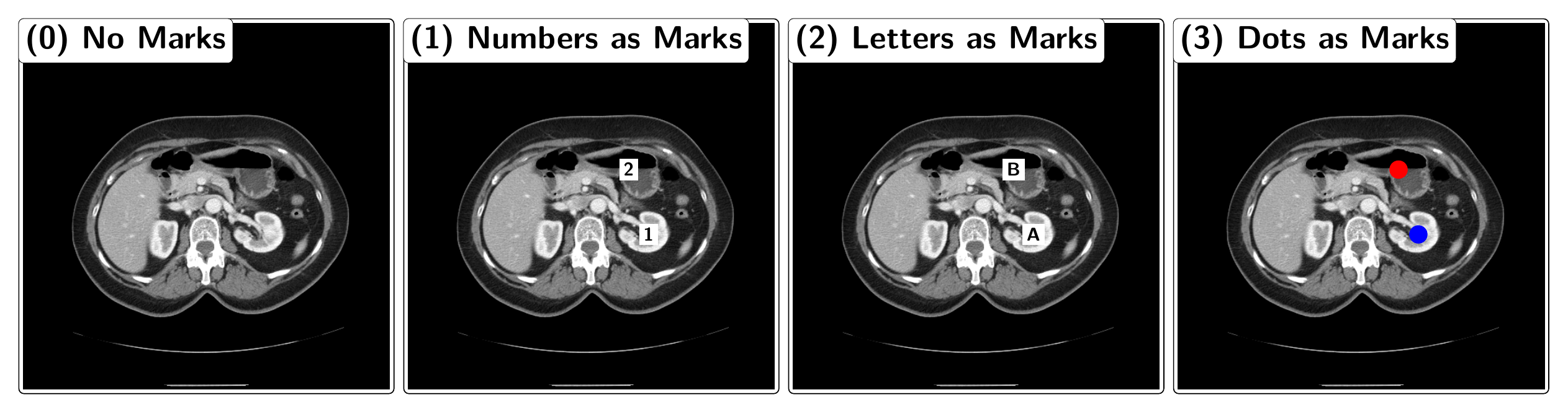}
    \caption{ 
    Different visual markers. 
    \textit{Note: The real marks are smaller as shown here.}}
    \label{fig:visprompt}
\end{figure}

We introduce the MIRP Benchmark dataset to systematically evaluate VLMs' ability to identify relative positions in medical images. 
The dataset consists of abdominal CT slices, each paired with a question about the relative position of two anatomical structures, following a standardized template:
\textit{``Is the {structure1} {above/below/to the left of/to the right of} the {structure2}?''}. The slices are randomly rotated and flipped to ensure that the model must derive relative positions from the image rather than relying on anatomical priors to produce a correct answer. The MIRP dataset is balanced, with an equal distribution of \emph{yes} and \emph{no} answers to questions. The two anatomical structures referenced in the question are optionally highlighted with visual markers. Three marker types are evaluated: (1) black numbers in a white box, (2) black letters in a white box, and (3) a red and a blue dot (see  Figure~\ref{fig:visprompt}). The following sections describe the dataset generation process and the VLM evaluation pipeline. While these sections are not required to understand the experiments and results, they provide additional context for our study.

\noindent \textbf{Dataset Generation:}
We combine CT volumes from two publicly available abdominal CT datasets: BTCV (Beyond the Cranial Vault)~\cite{landman2015miccai,gibson_2018_1169361} and AMOS (Abdominal Multi-Organ Segmentation)~\cite{ji2022amos,ji_yuanfeng_2022_7155725}. To extract anatomical structures from the volumetric data, we apply TotalSegmentator~\cite{wasserthal2023totalsegmentator}, a deep learning-based segmentation model built on nnU-Net~\cite{isensee2021nnu} and trained on diverse CT scans. TotalSegmentator provides segmentation masks for 117 anatomical structures, including organs, bones, muscles, and vessels. We extract axial PNG slices from the 3D volumes using SimpleITK~\cite{lowekamp2013design}. A slice is included if it contains at least two segmented anatomical structures, resulting in 4,878 image slices. Following consultation with senior radiologists, we use the soft tissue window as it best visualizes the segmented structures. For each 2D image, we randomly select two segmented anatomical structures that appear only once, are at least 50 pixels apart, and have a size at least twice that of the markers to generate question-answer pairs. The correct answer is determined based on the coordinates of the center of mass of the segmented structures. Visual markers (numbers, letters, dots) are also placed at the center of mass. We use the TotalSegmentator's structure numbers for all number marks (1-117) and map the letters to AA-EM, accordingly, so that each structure has a unique number and letter.\\
\textbf{VLM Evaluation Pipeline:}
We compare four different models, namely GPT-4o-2024-08-06 (OpenAI, closed-source, accessed via API), Pixtral-12B-2409~\cite{mistralpixtral} (Mistral, open-source), Llama3.2-11B-Vision-Instruct~\cite{metallama} (Meta, open-source) and JanusPro-7B~\cite{deepseekjanuspro} (DeepSeek-AI, open-source). All open-source models are accessed via HuggingFace. The prompts consist of a text prompt followed by the image. The text prompt contains a fixed and a variable component. The fixed component includes: Essential contextual information, indicating that the image represents an axial slice of an abdominal CT scan displayed in the soft tissue window; instructions to base the answer on the image rather than anatomy; an example response to illustrate the expected answer format. The variable text prompt component formulates the specific question.

\section{Experiments and Results}
\begin{figure}[h!]
    \centering
    \includegraphics[width=1\textwidth]{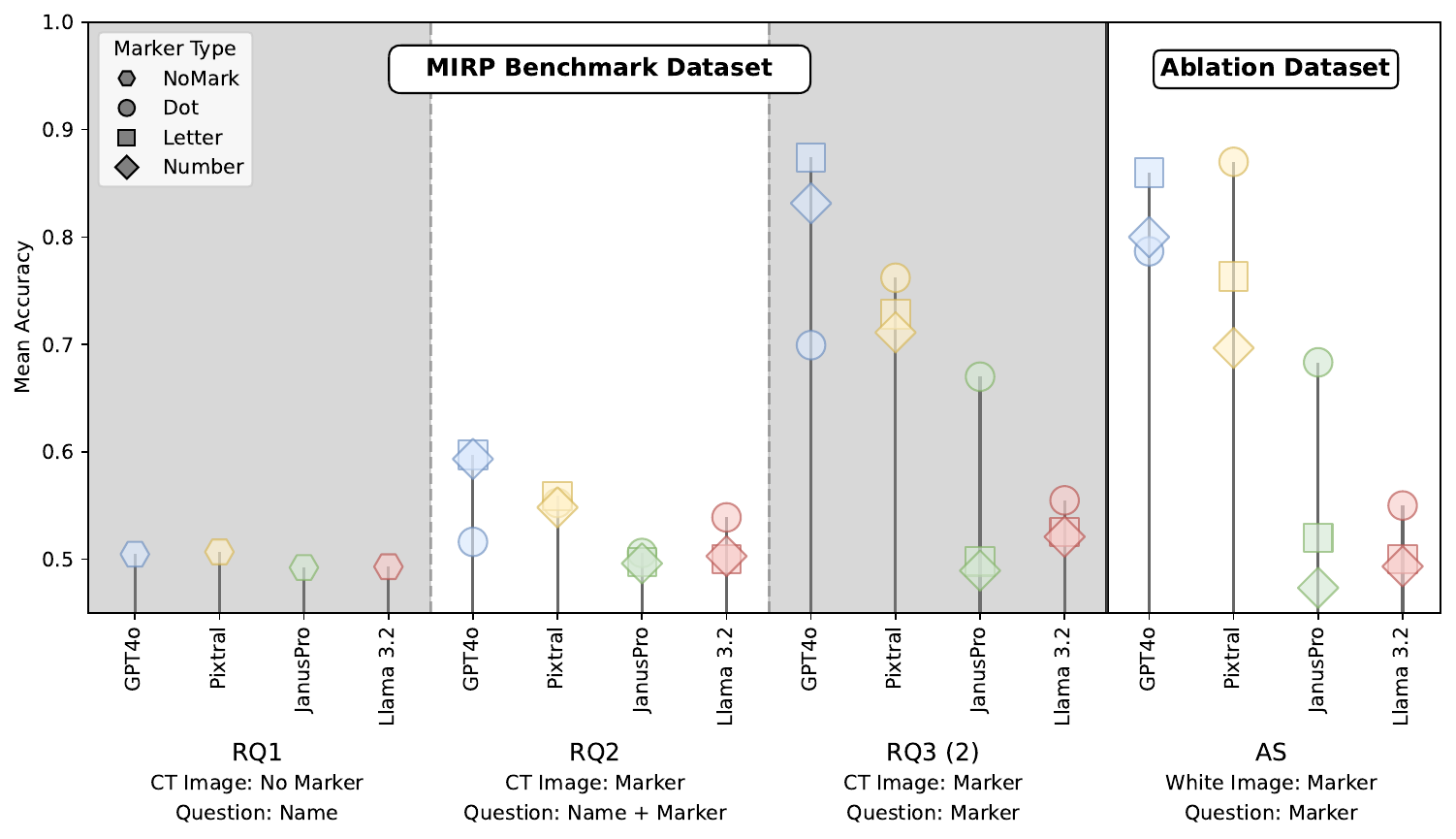}\textbf{}
    \caption{
    Mean accuracy for all experiments based on image-view evaluation on the MIRP benchmark dataset (RQ1-3) and the ablation dataset (AS). 
    }
    \label{fig:scatter}
\end{figure}

The following sections systematically investigate the three proposed research questions (RQ). Figure~\ref{fig:overview} illustrates an example of the VLM's input for each RQ. Each RQ is assessed through a binary classification task in which a model's response is considered correct if it matches the expected answer. Any other answer, including cases where the model fails to provide a valid response, is considered incorrect. Each experiment is repeated three times to report the mean and standard deviation. \\
\noindent\textbf{RQ1: Can current top-tier VLMs accurately determine relative positions in radiological images?}
To answer this, we test VLMs on plain rotated/flipped CT slices. We ask relative positioning questions following our question template, \eg{} \textit{``Is the left kidney below the stomach?''}. Figure~\ref{fig:scatter} and Table~\ref{tab:Results} show the results under ``RQ1''. All models reach accuracies that are around 50\%, which indicates that the models' performances are at chance level. This suggests an inability to determine relative positions from unmarked radiological images. \\
\noindent\textbf{RQ2: Can visual markers improve VLMs' ability to determine relative positions in radiological images?}
In this experiment, we use the images with additional markers (letters, numbers, red/blue dots). The template for the relative positioning questions is extended to include information about these markers, \eg{} \textit{``Is the left kidney (1) below the stomach (2)?''}, \textit{``Is the left kidney (A) below the stomach (B)?''} or \textit{``Is the left kidney (red) below the stomach (blue)?''}. Figure~\ref{fig:scatter} (RQ2) shows the results, with different objects referring to different visual markers (letters, numbers, dots). GPT-4o and Pixtral show slight performance improvements when letter or number markers are provided. However, JanusPro and Llama3.2 show little to no improvement. This indicates that visual markers alone may not be sufficient to enhance a VLM's ability to determine relative positions in radiological images.\\
\noindent\textbf{RQ3: Do VLMs prioritize prior anatomical knowledge over visual input when determining relative positions in radiological images?}
We hypothesize that VLMs do not rely solely on the provided image but instead base their answers to a large extent on learned anatomical knowledge within the language part. 
We test this hypothesis in two ways: (1) by assessing whether VLMs’ responses match the typical anatomical arrangement of the structures, which would require prior knowledge; and (2) by evaluating performance when anatomical names are removed from the prompt, thereby preventing the possibility of using prior anatomical knowledge.\\
\indent (1): Do VLMs' responses align with standard human anatomy?
Relative positions of anatomical structures can be provided in two ways: (a) based on the specific orientation of the provided rotated/flipped image, (b) based on the typical anatomical positions of the structures in standard human anatomy. So far, we have evaluated the VLMs' answers based on the image orientation. Below/above is defined only in the image, as its interpretation in anatomy is ambiguous (\ie{} anterior/posterior vs. superior/inferior). Left/right, however, is defined in both, image view and anatomically, allowing VLMs' responses to be evaluated based on either image-view correctness or anatomical correctness. For example, the correct answer to the question \textit{``Is the liver to the right of the stomach?''} would be yes based on standard human anatomy, since the liver is in most humans indeed to the right of the stomach (following the standard patient-oriented coordinate system used in radiology). But based on the provided image, the correct answer might be no. A yes response despite contradictory visual evidence suggests the model is relying on prior anatomical knowledge embedded in the language component rather than analyzing the image. To investigate what information (image vs. prior anatomy knowledge) a VLM bases its answer on, we use the same plain CT slices as in RQ1. We evaluate the VLMs' answers to the left/right questions in two ways: (a) correctness based on the provided rotated/flipped image (same evaluation as in RQ1) and (b) correctness based on how the structures are normally positioned in standard anatomy. The results based on the rotated/flipped image are: GPT-4o: 0.506; Pixtral: 0.508; JanusPro: 0.490; Llama3.2: 0.491. The results based on anatomical correctness are: GPT-4o: 0.757; Pixtral: 0.574; JanusPro: 0.493; Llama3.2: 0.547.
As already observed in RQ1, all VLMs perform at chance level when evaluated based on the image view. When evaluated for anatomical correctness, GPT-4o achieves an accuracy exceeding 75\%, demonstrating that most of its responses are consistent with the typical anatomical positions of the structures. As this information is not visually accessible in the rotated/flipped images, the model must rely on prior anatomical knowledge. Pixtral also performs better, but less markedly.
These results support our hypothesis that, for GPT-4o and Pixtral, responses are influenced more by prior anatomical knowledge than by image interpretation.
However, JanusPro and Llama3.2 remain at chance level, even when evaluated based on anatomical correctness.\\
\indent (2): How well do VLMs perform when prior knowledge cannot be used? We return to our primary goal, evaluating VLMs' performance based on the rotated/flipped image view. To do this, we conduct another experiment on the CT slices with markers, similar to RQ2. In RQ2, models were provided with the medical names of the anatomical structures in the text prompt questions. Now, we modify the questions so that models are given only visual markers with no reference to anatomical structures, \eg{} \textit{``Is the number 1 below the number 2?''}, \textit{``Is the letter A below the letter B?''}, or \textit{``Is the red dot below the blue dot?''}. An example input is shown in Figure~\ref{fig:overview} (RQ3). Now, the VLMs can no longer rely on prior anatomical knowledge based on the question and must instead base their answers entirely on the provided image. As shown in Figure~\ref{fig:scatter} under ``RQ3 (2)'', GPT-4o and Pixtral show a substantial increase in accuracy compared to RQ2. GPT-4o achieves over 85\% accuracy on letter markers, while Pixtral exceeds 75\% accuracy on dot markers. This indicates that these models can perform relative positioning tasks but struggle when anatomical terms are included in the question due to their learned priors. JanusPro and Llama3.2 also show improvements when using dot markers, but still lag behind GPT-4o
and Pixtral.\\
\indent These evaluations test our hypothesis that VLMs rely more on prior anatomical knowledge than on image content in two ways. First, when VLMs have access to anatomical names, GPT-4o and Pixtral produce more anatomically correct answers than image-based correct answers. This indicates a reliance on prior knowledge, as an anatomically correct answer can only stem from prior knowledge within the language part. Second, when anatomical names are removed, forcing the models to rely solely on image content, GPT-4o, and Pixtral achieve high accuracies when evaluated based on the image view. These findings support our hypothesis for GPT-4o and Pixtral but do not provide strong evidence for JanusPro and Llama3.2.\\
\noindent\textbf{Ablation Study: How well do VLMs perform on relative positioning tasks in general?}
To assess the ability to determine relative positions independent of domain knowledge, we design one of the simplest possible relative positioning tasks -- white images with randomly placed markers. We generate 100 images per marker type and ask questions such as \textit{``Is the number 1 above the number 2?''}. As shown in Fig.~\ref{fig:scatter} (AS), Pixtral demonstrates improved performance on dot markers, while the other models achieve results similar to those from RQ3. The fact that JanusPro and especially Llama3.2 do not perform well even under these simplified conditions suggests that both models have fundamental limitations in relative positioning tasks that extend beyond medical imaging.\\
\noindent\textbf{Further Analysis:}
Our findings across all experiments indicate that different models benefit from different visual markers. GPT-4o performs best with letter markers, while Pixtral, JanusPro, and Llama3.2 perform best with red/blue dots. Table~\ref{tab:Results} presents the results with the best-performing marker type per model for all experiments on image-view evaluation. Among the evaluated models, GPT-4o achieves the highest performance, whereas Pixtral is the best-performing open-source model. 
We further analyzed the results for each flip/rotation variant separately. Context: Radiological images in standard view mirror the anatomical definition (anatomical right appears on the image’s left). As a result, already in standard view, VLMs relying on prior anatomical knowledge rather than visual input answer left/right questions incorrectly. Since the left-right swap is a well-defined convention, VLMs might learn to compensate for it. To eliminate this potential bias, we applied the flips/rotations. Evaluation: In RQ1, for the two variants where the image aligns with anatomical left/right (Flip+NoRot and NoFlip+180°Rot), GPT-4o and Pixtral perform above chance, while for all other variants, performance is at chance level. This shows that VLMs do not compensate for the left-right swap, and it supports our RQ3 finding.

\begingroup
\setlength{\tabcolsep}{4.4pt}
\begin{table}[t]
    \caption{Accuracy for all experiments based on image-view evaluation. 
    For RQ2, RQ3, and AS, the best-performing markers are reported: letters for GPT-4o; dots for Pixtral, JanusPro, and Llama3.4.
    }
    \centering
    \resizebox{\textwidth}{!}{
    \fontsize{8}{9}\selectfont
    \begin{tabular}{lcccc}
        \toprule
        Model & RQ 1 & RQ 2 & RQ 3 (2) & \color{gray} AS  \\                     
        \midrule
      GPT-4o  &  0.505  $\pm$ 0.005   &  0.597  $\pm$ 0.004  & 0.874  $\pm$ 0.002 & \color{gray} 0.860 $\pm$ 0.026 \\
      Pixtral  &  0.507  $\pm$ 0.006   &  0.552  $\pm$ 0.003  & 0.762  $\pm$ 0.004 & \color{gray} 0.870 $\pm$ 0.010  \\
      JanusPro  &  0.492  $\pm$ 0.008   &  0.506  $\pm$ 0.006  & 0.670  $\pm$ 0.007 & \color{gray} 0.683 $\pm$ 0.023 \\
      Llama3.2  &  0.493 $\pm$ 0.005  &  0.539  $\pm$ 0.008  & 0.555  $\pm$ 0.007 & \color{gray} 0.550 $\pm$ 0.053  \\
        \bottomrule
    \end{tabular}
    }
    \label{tab:Results}
\end{table}
\endgroup

\section{Conclusion and Future Work}
Support from a VLM in a radiology department could drastically improve the efficiency and accuracy of diagnostics. However, this vision can only be realized if VLMs can accurately answer relative positioning questions. Our study demonstrates that current VLMs fail at this fundamental task, performing at chance level when provided with unmarked CT slices. When introducing 
markers, we achieve moderate improvements, but results remain far below those observed in natural image studies~\cite{yang2023set,cai2024vip}. Since VLMs already possess strong anatomical priors embedded in the language components~\cite{tarabanis2024performance}, we hypothesize that models rely more on prior knowledge than on actual image content. This bias could result in critical misdiagnoses, particularly in anatomical anomalies or rare conditions. Our findings support this hypothesis for GPT-4o and Pixtral, whereas we do not find strong supporting evidence for JanusPro and Llama3.2. A task with no medical context confirms that JanusPro and Llama3.2 struggle with identifying relative positions in general. Altogether, further research that builds on these findings is essential for implementing VLM support in radiology. To facilitate such research, we introduce MIRP, a publicly available benchmark dataset, alongside this evaluation.
While we evaluate general-purpose VLMs on the MIRP benchmark, future research could assess medical fine-tuned VLMs~\cite{li2023llava,hartsock2024vision}. Our study and the MIRP benchmark dataset are limited to CT images. While we assume that the presented findings generalize to other modalities, future work could explore MRI or PET images. Furthermore, our evaluation is based on 2D slices rather than full 3D volumes since current VLMs operate on 2D images. However, transitioning to volumetric data could drastically improve clinical applicability. Altogether, our evaluation of the ability to identify relative positions in medical images serves as a critical first step toward enabling VLMs for clinical applications and paves the way for myriad follow-up experiments that are required before bringing VLMs into real-world radiology practice.

\begin{credits}
\subsubsection{\ackname}  This study was funded by the German Ministry of Education and Research as part of the University Medicine Network (Project: RACOON, 01KX2121) and by the German Research Foundation DFG (Project: KEMAI, GRK 3012 – 520750254).

\subsubsection{\discintname}
The authors have no competing interests to declare that are relevant to the content of this article.
\end{credits}

\bibliographystyle{splncs04}
\bibliography{mybibliography}

\begin{thebibliography}{10}
\providecommand{\url}[1]{\texttt{#1}}
\providecommand{\urlprefix}{URL }
\providecommand{\doi}[1]{https://doi.org/#1}

\bibitem{agolia2019preventing}
Agolia, J.P., et~al.: Preventing wrong-level spine surgery. In: International Conference on Complications in Neurosurgery. pp.~1--8. Springer (2019)

\bibitem{agrawal2024pixtral}
Agrawal, P., et~al.: Pixtral 12b. arXiv preprint arXiv:2410.07073  (2024)

\bibitem{almeida2023performance}
Almeida, L.C., et~al.: Performance of {ChatGPT} on the {B}razilian radiology and diagnostic imaging and mammography board examinations. Radiology: Artificial Intelligence  \textbf{6}(1),  e230103 (2023)

\bibitem{bae2024mimic}
Bae, S., et~al.: {MIMIC}-{Ext}-{MIMIC}-{CXR}-{VQA}: {A} complex, diverse, and large-scale visual question answering dataset for chest {X}-ray images. PhysioNet  (2024)

\bibitem{brady2017error}
Brady, A.P.: Error and discrepancy in radiology: inevitable or avoidable? Insights into imaging  \textbf{8},  171--182 (2017)

\bibitem{cai2024vip}
Cai, M., et~al.: {ViP}-{LLaVA}: Making large multimodal models understand arbitrary visual prompts. In: Proceedings of the IEEE/CVF Conference on Computer Vision and Pattern Recognition. pp. 12914--12923 (2024)

\bibitem{chen2024quantifying}
Chen, M., et~al.: Quantifying and mitigating unimodal biases in multimodal large language models: {A} causal perspective. arXiv preprint arXiv:2403.18346  (2024)

\bibitem{chen2025janus}
Chen, X., et~al.: Janus-pro: Unified multimodal understanding and generation with data and model scaling. arXiv preprint arXiv:2501.17811  (2025)

\bibitem{joint2025sentinel}
Commission, J., et~al.: Sentinel event data 2023 annual review. (2025)

\bibitem{datta2020understanding}
Datta, S., et~al.: Understanding spatial language in radiology: {R}epresentation framework, annotation, and spatial relation extraction from chest {X}-ray reports using deep learning. Journal of biomedical informatics  \textbf{108},  103473 (2020)

\bibitem{deepseekjanuspro}
DeepSeek-AI: Janus-pro-7b [acsession date: 2025-06-01] \url{https://huggingface.co/deepseek-ai/Janus-Pro-7B}

\bibitem{dubey2024llama}
Dubey, A., et~al.: The llama 3 herd of models. arXiv preprint arXiv:2407.21783  (2024)

\bibitem{eitler2022situs}
Eitler, K., et~al.: Situs inversus totalis: a clinical review. International journal of general medicine pp. 2437--2449 (2022)

\bibitem{european2011good}
ESR: Good practice for radiological reporting. {G}uidelines from the {E}uropean society of radiology {(ESR)}. Insights into imaging  \textbf{2}(2),  93--96 (2011)

\bibitem{gao2025gllava}
Gao, J., et~al.: G-{LL}a{VA}: Solving geometric problem with multi-modal large language model. In: The Thirteenth International Conference on Learning Representations (2025)

\bibitem{gibson_2018_1169361}
Gibson, E., et~al.: Multi-organ abdominal {CT} {R}eference {S}tandard {S}egmentations, \url{https://doi.org/10.5281/zenodo.1169361}

\bibitem{hartsock2024vision}
Hartsock, I., Rasool, G.: Vision-language models for medical report generation and visual question answering: {A} review. Frontiers in Artificial Intelligence  \textbf{7},  1430984 (2024)

\bibitem{hurst2024gpt}
Hurst, A., et~al.: Gpt-4o system card. arXiv preprint arXiv:2410.21276  (2024)

\bibitem{isensee2021nnu}
Isensee, F., et~al.: nn{U}-net: a self-configuring method for deep learning-based biomedical image segmentation. Nature methods  \textbf{18}(2),  203--211 (2021)

\bibitem{ji2022amos}
Ji, Y., et~al.: Amos: {A} large-scale abdominal multi-organ benchmark for versatile medical image segmentation. Advances in neural information processing systems  \textbf{35},  36722--36732 (2022)

\bibitem{jin2024hidden}
Jin, Q., et~al.: Hidden flaws behind expert-level accuracy of multimodal {GPT}-4 vision in medicine. npj Digital Medicine  \textbf{7}(1), ~190 (2024)

\bibitem{kanal2017us}
Kanal, K.M., et~al.: {US} diagnostic reference levels and achievable doses for 10 adult {CT} examinations. Radiology  \textbf{284}(1),  120--133 (2017)

\bibitem{kang2021factors}
Kang, J.D., et~al.: Factors associated with missed and misinterpreted cases of pancreatic ductal adenocarcinoma. European Radiology  \textbf{31},  2422--2432 (2021)

\bibitem{kim2002postoperative}
Kim, K.W., et~al.: Postoperative anatomic and pathologic findings at {CT} following gastrectomy. Radiographics  \textbf{22}(2),  323--336 (2002)

\bibitem{landman2015miccai}
Landman, B., et~al.: Miccai multi-atlas labeling beyond the cranial vault--workshop and challenge. In: Proc. MICCAI Multi-Atlas Labeling Beyond Cranial Vault—Workshop Challenge. vol.~5, p.~12 (2015)

\bibitem{lau2018dataset}
Lau, J.J., et~al.: A dataset of clinically generated visual questions and answers about radiology images. Scientific data  \textbf{5}(1),  1--10 (2018)

\bibitem{li2023llava}
Li, C., et~al.: Llava-med: {T}raining a large language-and-vision assistant for biomedicine in one day. Advances in Neural Information Processing Systems  \textbf{36},  28541--28564 (2023)

\bibitem{lowekamp2013design}
Lowekamp, B.C., et~al.: The design of {SimpleITK}. Frontiers in neuroinformatics  \textbf{7}, ~45 (2013)

\bibitem{marcu2022impact}
Marcu, D., et~al.: Impact of retroperitoneal tumors on the digestive tract. Experimental and Therapeutic Medicine  \textbf{24}(6), ~1--8 (2022)

\bibitem{metallama}
Meta: Llama-3.2-11b-vision-instruct [acsession date: 2025-06-01] \url{https://huggingface.co/meta-llama/Llama-3.2-11B-Vision-Instruct}

\bibitem{mistralpixtral}
Mistral: Pixtral-12b-2409 [acsession date: 2025-06-01] \url{https://huggingface.co/mistralai/Pixtral-12B-2409}

\bibitem{nan2024beyond}
Nan, Y., et~al.: Beyond the hype: {A} dispassionate look at vision-language models in medical scenario. arXiv preprint arXiv:2408.08704  (2024)

\bibitem{rahmanzadehgervi2024vision}
Rahmanzadehgervi, P., Bolton, L., Taesiri, M.R., Nguyen, A.T.: Vision language models are blind. In: Proceedings of the Asian Conference on Computer Vision. pp. 18--34 (2024)

\bibitem{srivastav2023chatgpt}
Srivastav, S., et~al.: {ChatGPT} in radiology: the advantages and limitations of artificial intelligence for medical imaging diagnosis. Cureus  \textbf{15}(7) (2023)

\bibitem{tarabanis2024performance}
Tarabanis, C., et~al.: Performance of publicly available large language models on internal medicine board-style questions. PLOS Digital Health  \textbf{3}(9) (2024)

\bibitem{vo2025vision}
Vo, A., Nguyen, K.N., Taesiri, M.R., Dang, V.T., Nguyen, A.T., Kim, D.: Vision language models are biased. arXiv preprint arXiv:2505.23941  (2025)

\bibitem{waite2017interpretive}
Waite, S., et~al.: Interpretive error in radiology. American Journal of Roentgenology  \textbf{208}(4),  739--749 (2017)

\bibitem{wasserthal2023totalsegmentator}
Wasserthal, J., et~al.: Total{S}egmentator: {R}obust segmentation of 104 anatomic structures in {CT} images. Radiology: Artificial Intelligence  \textbf{5}(5) (2023)

\bibitem{yang2023set}
Yang, J., et~al.: Set-of-mark prompting unleashes extraordinary visual grounding in gpt-4v. arXiv preprint arXiv:2310.11441  (2023)

\bibitem{ji_yuanfeng_2022_7155725}
Yuanfeng, J.: Amos: {A} large-scale abdominal multi-organ benchmark for versatile medical image segmentation (Nov 2022). \doi{10.5281/zenodo.7155725}

\end{thebibliography}

\end{document}